\documentclass[lettersize,journal]{IEEEtran}

\usepackage{amsmath,amsfonts}
\usepackage{algorithmic}
\usepackage{algorithm}
\usepackage{array}
\usepackage[justification=centering]{caption}
\usepackage[caption=false,font=normalsize,labelfont=sf,textfont=sf]{subfig}
\usepackage{textcomp}
\usepackage{stfloats}
\usepackage{url}
\usepackage{verbatim}
\usepackage{graphicx}
\usepackage{cite}
\usepackage{multirow}
\usepackage{makecell}
\usepackage[english]{babel}
\usepackage{booktabs}
\usepackage[hidelinks]{hyperref}
\usepackage{pifont}
\usepackage{xcolor}
\usepackage{fontawesome}

\usepackage{multirow} 

\usepackage{algorithm}
\usepackage{algorithmic}
\usepackage{enumitem}

\usepackage[english]{babel}

\usepackage[letterpaper,top=2cm,bottom=2cm,left=3cm,right=3cm,marginparwidth=1.75cm]{geometry}

\usepackage{amsmath}
\usepackage{graphicx}

\begin{document}
\title{MASR: Self-Reflective Reasoning through Multimodal Hierarchical Attention Focusing for Agent-based Video Understanding}

\DeclareRobustCommand*{\IEEEauthorrefmark}[1]{%
	\raisebox{0pt}[0pt][0pt]{\textsuperscript{\footnotesize #1}}%
}

\author{
	\IEEEauthorblockN{
		Shiwen Cao\IEEEauthorrefmark{1},
		Zhaoxing Zhang\IEEEauthorrefmark{1},
		Junming Jiao\IEEEauthorrefmark{1},
		Juyi Qiao\IEEEauthorrefmark{1},
		Guowen Song\IEEEauthorrefmark{1},
        Rong Shen\IEEEauthorrefmark{1}$^{\ast}$ \thanks{$^{\ast}$Corresponding to: shenrong@lixiang.com}
		Xiangbing Meng\IEEEauthorrefmark{2},
	}\\
	\IEEEauthorblockA{
		\IEEEauthorrefmark{1}Li Auto Inc., Beijing, China\\
		\IEEEauthorrefmark{2}University of Chinese Academy of Sciences, Beijing, China\\
	}
}

\maketitle
\begin{abstract}
Even in the era of rapid advances in large models, video understanding remains a highly challenging task. Compared to texts or images, videos commonly contain more information with redundancy, requiring large models to properly allocate attention at a global level for comprehensive and accurate understanding. To address this, we propose a Multimodal hierarchical Attention focusing Self-reflective Reasoning (MASR) framework for agent-based video understanding. The key innovation lies in its ability to detect and prioritize segments of videos that are highly relevant to the query. Firstly, MASR realizes Multimodal Coarse-to-fine Relevance Sensing (MCRS) which enhances the correlation between the acquired contextual information and the query. Secondly, MASR employs Dilated Temporal Expansion (DTE) to mitigate the risk of missing crucial details when extracting semantic information from the focused frames selected through MCRS. By iteratively applying MCRS and DTE in the self-reflective reasoning process, MASR is able to adaptively adjust the attention to extract highly query-relevant context and therefore improve the response accuracy. In the EgoSchema dataset, MASR achieves a remarkable 5\% performance gain over previous leading approaches. In the Next-QA and IntentQA datasets, it outperforms the state-of-the-art standards by 0.2\% and 0.3\% respectively. In the Video-MME dataset that contains long-term videos, MASR also performs better than other agent-based methods.
\end{abstract}

\section{Introduction}
In recent years, video has been increasingly utilized in various domains, due to its ability to convey more extensive information. Consequently, video-understanding tasks have progressively become a hotspot in the multimodal research field.
Compared with text and image data, video data span broader both spatially and temporally, exhibit more complex semantic contents with multiple modalities, feature strong causal relationships with redundancy, all of which pose great challenges to the video analysis methods.

With the remarkable success of Large Language Models (LLMs) [1]-[8], [70], [71] in natural language processing, Multimodal Large Language Models (MLLMs) [9]-[21], [74], [75] established on LLMs have demonstrated impressive capabilities across various image understanding tasks, including recognition [22], object detection [14], [15] and vision navigation [23]. However, applying MLLM-centric approaches to video understanding tasks remains difficult in view of the following aspects [24], [50], [57], [59]:
\begin{itemize}[leftmargin=*]
\item 
\textbf{Larger Data Volume:} As video quality improves, higher resolutions and frame rates lead to massive data that pose significant challenges for the MLLMs to process.

\item 
\textbf{Higher Information Density:} Long videos often contain multiple scenes with rich semantic expressions. Pure compression or downsampling often fails to preserve fine-grained information, resulting in incorrect responses to detail-oriented questions.

\item
\textbf{Higher Redundancy:} Videos commonly exhibit temporal redundancy. Without precise identification and proper selection of key frames, the attention of MLLMs will be placed upon these portions that are not closely connected to the query and may suffer misinterpretations due to this comprehension bias.
\end{itemize}

Consequently, a video understanding framework is needed that enables MLLMs to selectively absorb multimodal video information and adaptively focus on different parts of the video to get the well-rounded answer iteratively like the human's thinking process.

Some video understanding solutions leverage pre-trained video encoders in combination with task-specific, transformer-based multimodal fusion manipulations. The spatio-temporal tokens produced by these modules are then fed into MLLMs for comprehension. Such MLLM-based video understanding methods (video-MLLM) [25]-[42] have achieved favorable results on certain datasets. However, they often require task-specific supervised fine-tuning, which limit their generalization capabilities. Moreover, when confronted with long videos with multiple scenes and rich details, the global narrative logic and fine-grained but critical information become obscured within the excessive data volume, making them hard for MLLMs to sense. Furthermore, the compression of visual tokens is inevitable due to the context length constraints that further destroys the fine-grained information perception capability of MLLMs.

In contrast to Video-MLLM-based approaches, agent-based frameworks leverage the comprehension and decision-making capabilities of pre-trained language models to construct multi-agent collaborative systems that simulate human cognitive mechanisms. These frameworks demonstrate unique advantages in dynamic task allocation and automated tool-use capabilities [43]. As illustrated in Figure 1, current mainstream agent-based video understanding methods employ a systematic answer-evaluate-refine architecture. By harnessing the self-reflective reasoning abilities of LLMs or MLLMs, they adopt a divide-and-conquer approach to video comprehension, enabling solutions based on such architectures [44]-[60] to exhibit superior generalization and adaptability with comparable core model sizes. Furthermore, the training-free nature of these frameworks significantly reduces dependence on high-quality annotated data.

Therefore, we propose the \textbf{M}ultimodal and hierarchical \textbf{A}ttention focusing \textbf{S}elf-Reflective \textbf{R}easoning (MASR) framework, an agent-based video understanding solution that imitates human cognitive strategies for video understanding. MASR is able to dynamically adjust its attention based on feedback from previous contemplation. Inheriting the advantages of agent-based frameworks mentioned above, MASR also owns the capability to integrate various mainstream LLMs and MLLMs to enable efficient video understanding.

The core innovative works of our MASR lies in its ability to precisely sense and prioritize query-relevant segments from large volume of video data (highlighted in the lavender regions of Figure \ref{MASR1}). In summary, these key contributions are listed as follows:

\begin{figure}
\centering
\includegraphics[width=1.0\linewidth]{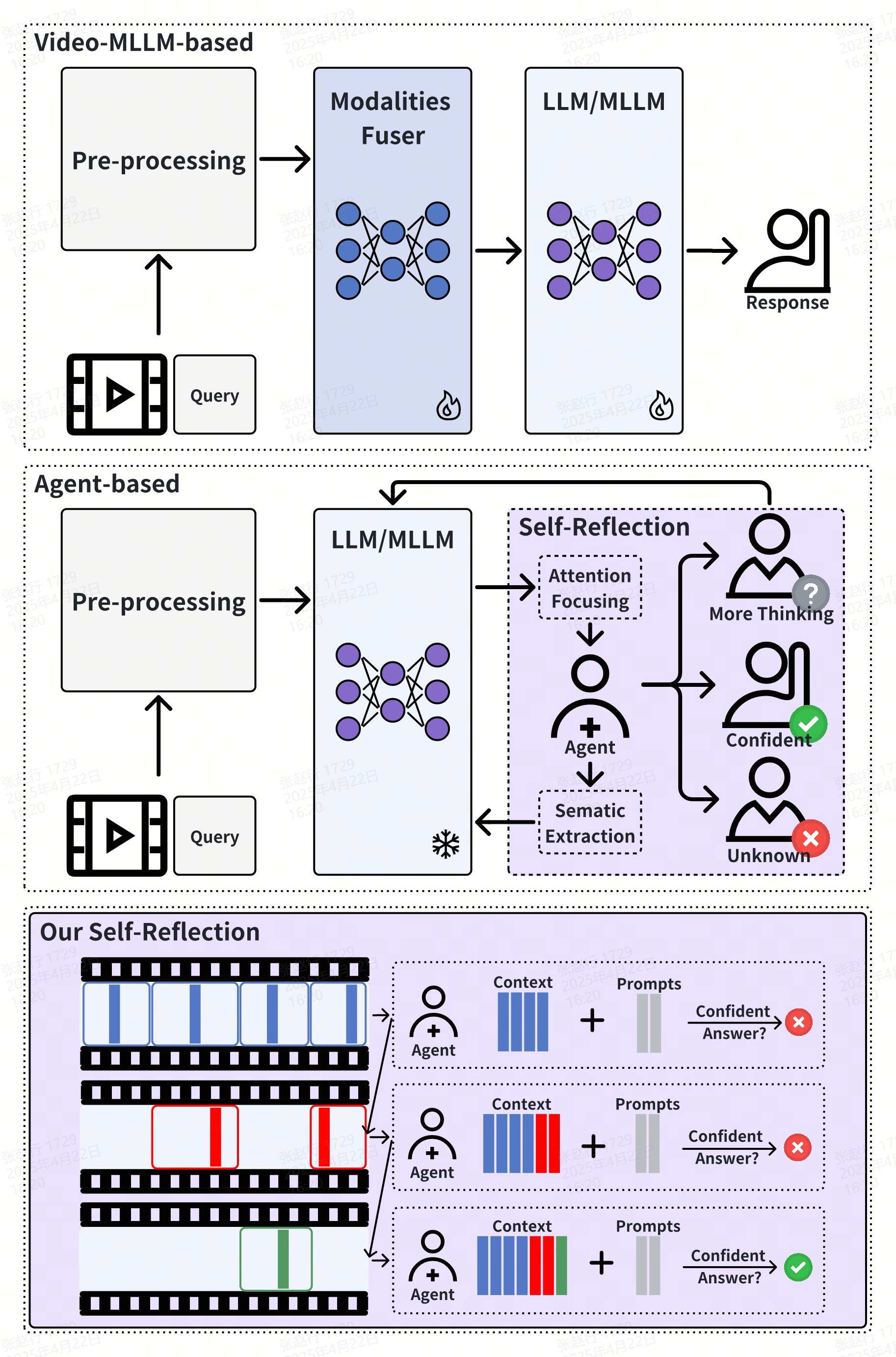}
\captionsetup{justification=raggedright, singlelinecheck=false}
\caption{\label{MASR1}A comparison of the two mainstream MLLM-based video understanding frameworks: Video-MLLM-based and Agent-based. In particular, the purple-highlighted sections in the agent-based method indicate the creative works in our MASR framework.}
\end{figure}

\begin{itemize}[leftmargin=*]
\item
\textbf{Multimodal hierarchical relevance retrieval with spatio-temporal enhancement:} We develop a novel multimodal-based hierarchical relevance filtering module, coupled with an efficient semantic extraction policy in order to retrieve the most query-relevant context for the LLM to focus, thus improving both the effectiveness and comprehensiveness of long-form video understanding.
\item
\textbf{Efficient self-reflection mechanism:} We implement an adaptive self-reflection mechanism using a single LLM. Through iterative attention adaption guided by response confidence feedback, the system autonomously acquires high-relevance contextual information, achieving measurable accuracy improvements. Comparison experiments on the main video QA benchmarks demonstrate the superiority of our single LLM-based self-reflection module.
\item
\textbf{Plug-and-play architecture:} MASR is compatible with concurrent mainstream LLMs and MLLMs. Its architecture ensures that our solution's performance automatically benefits from future advances in these LLMs and MLLMs.
\end{itemize}

\section{Related Works}
With the rapid advancement of MLLM technology, state-of-the-art (SOTA) frameworks in video understanding commonly employ LLMs or MLLMs in recent years. These large models can be generally categorized into the following two approaches:

\begin{itemize}[leftmargin=*]
\item
\textbf{Video-MLLM-based:} These solutions usually take advantage of the tokenized outputs from MLLMs as hidden states, thus embedding their pre-trained general multimodal understanding capabilities into task-specific network architectures for video comprehension. Through supervised fine-tuning the parameters in the large models or adapters [25]-[42] using in-domain video datasets, these frameworks are able to give precise response for the video understanding.

However, these video-MLLM-based methods typically exhibit several inherent limitations: First of all, their overall architectures tend to be complex and need to go through massive training at the cost of huge data curation and annotation. Secondly, supervised fine-tuning procedures often weaken the models' generalization ability. Again, the indispensable data compression manipulations such as token compression and temporal sliding windows frequently degrade understanding accuracy owing to the loss of details. Lastly, these systems generally fall short in terms of self-directed exploration, which are essential for tackling complex tasks such as video understanding [24], [50], [57], [59].
\item
\textbf{Agent-based:} This types of approaches [44]-[61] commonly leverage the scheduling capability for pre-trained LLMs or MLLMs to help video understanding. By efficiently integration with self-reflective close-loop, the general comprehension abilities in these LLMs or MLLMs in agent-based framework can be maximized in video understanding tasks without compromising their general comprehension capabilities. The agent-based methods also allow us to feet the LLMs or MLLMs with the selected contents through other manipulations to enhance the comprehension efficiency in the same core model configuration.

Among the other mainstream SOTA frameworks, DrVideo [49] employs an iterative self-reflection mechanism to pinpoint relevant video segments for comprehension. However, it relies entirely on LLMs to assess relevance through text-converted image modalities, leading to comprehension errors due to insufficient utilization of rich visual information. VideoTree [59] absorbs visual features in its self-reflective reasoning, yet only forms a closed loop for partial attention allocation processes, lacking dynamic adjustment of allocation policies based on final answer quality. Its adaptive breadth expansion mechanism is also more complex than our \textbf{D}ilated \textbf{T}emporal \textbf{E}xpansion (DTE) policy. VCA [50] introduces additional MLLM-based evaluation models to adaptively adjust focus clips through multimodal information, but suffers from performance degrading due to excessive model involvement, with its core attention focusing effectiveness being limited by historical image storage capacity. Both VideoAgent [57] and VideoINSTA [60] leverage auxiliary models to extract auxiliary information (e.g., foreground object locations) for video understanding. However, the introduction of these auxiliary models exhibit heavy computational overhead. They also fail to establish effective closed-loop optimization through response evaluation.
\item
\textbf{Our solution:} Drawing inspiration from these examples above, we propose the novel MASR. Following human cognitive processes to understand video, it significantly improves the video understanding accuracy through multimodal hierarchical attention focusing manipulation both spatially and temporally. Compared to its concurrent solutions, MASR first introduces an enhanced context retrieval mechanism that ensures both relevance and completeness of the acquired context for the responding model. It then incorporates this creative retrieval mechanism into a self-reflective reasoning mechanism guided by answer confidence feedback using a single LLM. This efficient realization shows superior performance across both long-form such as Video-MME [67] and medium-short-form like EgoSchema [64], Next-QA [65], and IntentQA [66] video question answering benchmarks.
\end{itemize}

\section{Methodology}
Our proposed MASR mirrors the way humans reason through video-based questions by arranging its modules in the following way:

\begin{itemize}[leftmargin=*]
\item
Perform a quick end-to-end scan of the video and decompose it into relatively independent semantic segments.
\item
Coarsely predict which semantic segments is highly possible to contain the necessary information.
\item
Conduct a finer-grained focus upon these segment candidates from to acquire a well-rounded context with informative details.
\item
Dynamically evaluates whether the collected context is sufficient and repeat the last two steps to adjust the emphasis if current contextual information is not enough to generate a confident response.
\end{itemize}

\begin{figure*}
\centering
\includegraphics[width=1.0\linewidth]{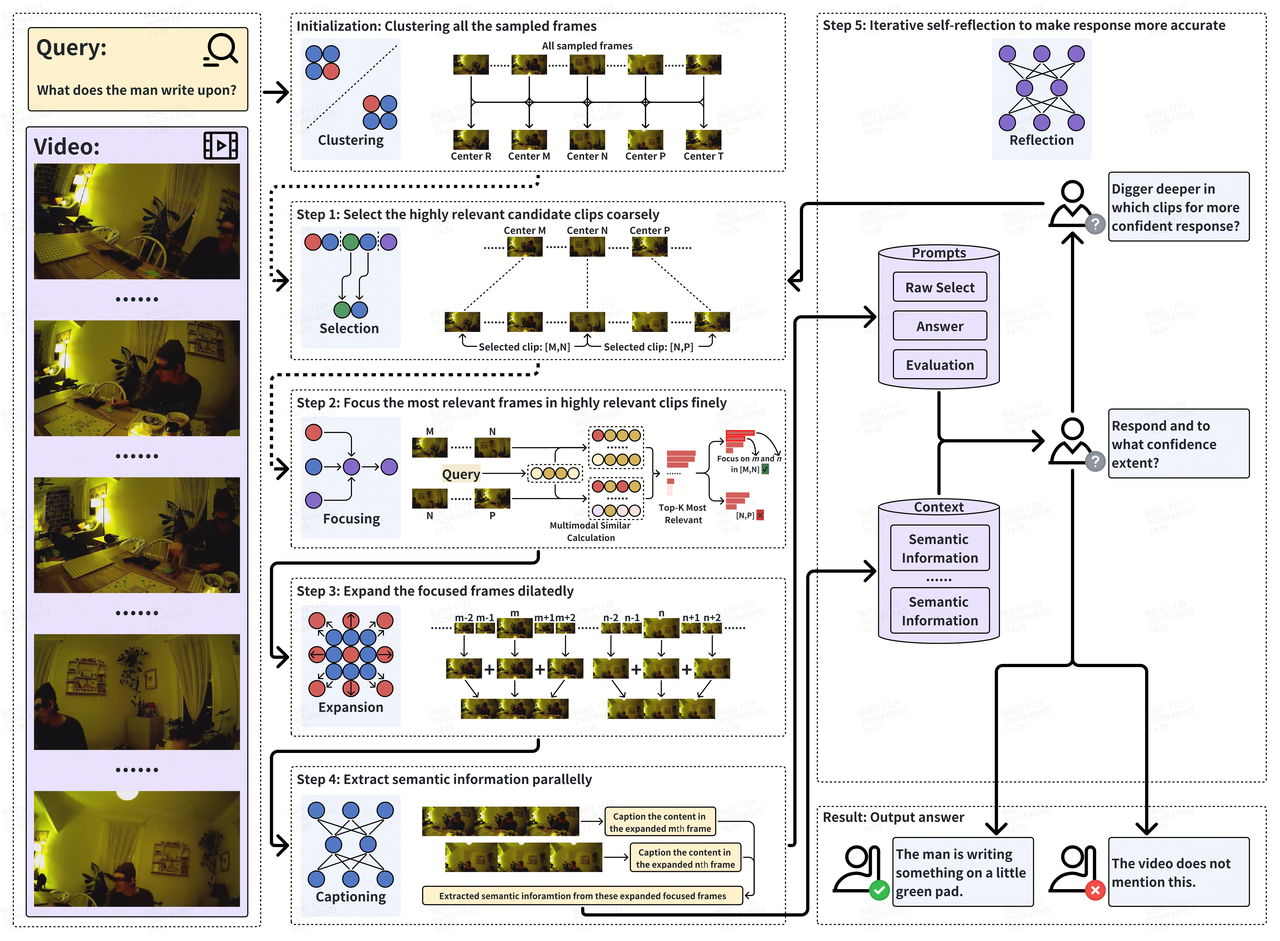}
\captionsetup{justification=raggedright, singlelinecheck=false}
\caption{\label{MASR2}A illustration of the complete MASR pipeline, where the leftmost section displays the input query and all video frames and the other sections shows core modules: Step 1 employs "selection" to denote the coarse attention focusing process that identifies highly relevant clip candidates based on semantic features; Step 2 utilizes "focusing" to represent the fine attention focusing process that pinpoints highly relevant frames through semantic-visual feature similarity matching; Step 3 performs DTE on selected frames; Step 4 extracts semantic features from expanded frames via VLM as contextual information for question answering; Step 5 generates responses while evaluating confidence scores to determine whether to output directly or reiterate the focusing-selection process for missing information. Notably, a single LLM in Step 5 serves as a reflector for response generation, confidence evaluation, and Step 1's coarse attention focusing. Since the context for initial coarse focusing remain empty during the first self-reflection round, MASR directly input clustered center frames obtained in the initialization stage as highly relevant frames to Step 3. This is indicated by dashed lines in the diagram.}
\end{figure*}

In our pipeline, the MASR begins by performing video clip-wise clustering on the input video. Through multi-level and multimodal relevance assessment, the MASR recognizes the most query-relevant frames and imposes the DTE upon these frames to widen the comprehension horizon. A Vision-Language Model (VLM) subsequently extracts semantic information from these highly relevant frames after DTE as the focused contextual basis for answering. The framework employs answer confidence scores as feedback to iteratively adjust its multi-level and multimodal relevance assessment until a confident enough response is acquired. Figure \ref{MASR2} illustrates how MASR dynamically balances efficiency and precision through its multimodal hierarchical attention focusing mechanism.

The attention focusing mechanism in MASR consists of these key components: 
\begin{itemize}[leftmargin=*]
\item
Video clip-wise clustering based on visual features during initialization.
\item
\textbf{M}utilmodal \textbf{C}oarse-to-fine \textbf{R}elevance \textbf{S}ensing (MCRS) capability comprises of LLM-assisted coarse selection of semantically highly query-relevant video clips combined with multimodal fine-grained relevant frame sensing within the scopes of previous coarse-selected relevant clips. 
\item
DTE manipulation of the focused query-relevant frames to widen the temporal receptive field while preserving critical information.
\item
Adjust attention focusing hierarchically and iteratively through responding confidence self-reflection utilizing a single LLM.
\end{itemize}

\subsection{\textbf{Video Clip-wise Clustering}}
Given a video $V \in R^{C\times{H}\times{W}\times{T}}$ comprising $T$ image frames 
and a textual query $Q$, we first perform frame sampling (e.g., uniform sampling) on the video to obtain the sampled frame set $spl\_frms=\{F_i\}_{i=1}^{T/t}$, where the sampling interval $t$ is determined by the total frame count $T$ and image resolution to ensure that a decent number of frames are selected for subsequent processing. $spl\_frms$ then undergo visual feature-based clustering to produce $N$ cluster center frames and $N+1$ video clips segmented by these center frames as demonstrated in the initialization phase.
\subsection{\textbf{Mutilmodal Coarse-to-fine Relevance Sensing (MCRS)}}

\textbf{Coarse Selection:} Since the context is short of pre-known knowledge about the target video after the clustering in the initialization stage, the LLM in the MASR fails to perform the coarse selection and we include all clustered center frames as highly query-relevant frames to initiate the self-reflection process. As the context is updated in the subsequent self-reflection rounds, the LLM owns the coarse selection capability and gives the raw selection results as the Step 1 in Figure \ref{MASR2} shows.

\textbf{Fine Focusing:} According to the coarse selection policy, these selected video clip candidates retain the average relevance with the query. It is possible that they miss critical details under the circumstances that query-related content is presented visually (e.g., rapid foreground position changes within short durations). To address this, our MASR introduces an additional visual feature-based relevance screening mechanism that refines the focus granularity to the frame level through token-based similarity matching.

\begin{figure*}
\centering
\includegraphics[width=1.0\linewidth]{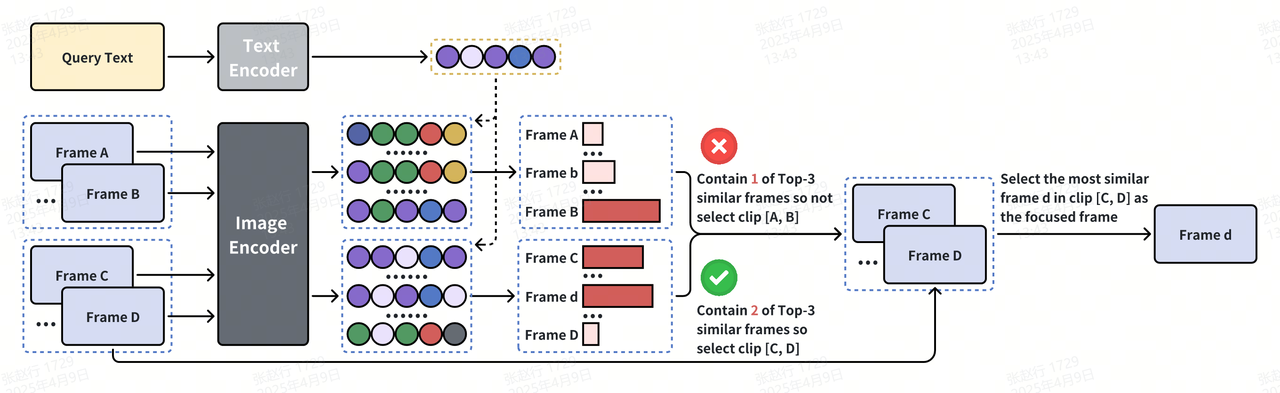}
\captionsetup{justification=raggedright, singlelinecheck=false}
\caption{\label{MASR3}A example of the implementation of fine-grained relevance sensing process. Between the two coarsely-selected video clip candidates [A, B] and [C, D], our fine-focusing algorithm determines that the
latter segment exhibits higher relevance to the query, as it contains more relevant visual tokens.
Consequently, we select frame d, which has the highest similarity score in the video clip [C, D].}
\end{figure*}

Specifically, we first encode the $K_c$ frames from updated candidate video clip set $spl\_clps$ using an image encoder. For simplicity, we assume all modality-encoded features are in the same dimension $d$, yielding visual tokens $Tk_{v\_{cf}}\in R^{{N_{cf}}\times{d}}$ for all the compared frames. These visual tokens are compared with the query's text token $Tk_{t}\in R^{1\times d}$
via cosine similarity computation and sorting $Sim_{v\_cf}=sort(Tk_{v\_cf} \cdot (Tk_t)^{T})\in R^{N_{cf}}$ to generate frame-wise similarity scores respectively. The top-$K_v$ visual tokens are considered as highly query-relevant within each of these candidates. We then count each candidate clip's number of such relevant tokens, select the top-$K_f$ clips with largest number of counts, and retrieve their most query-similar frames as fine-focused relevant frames to $fcs\_frms$. The whole process is simply visualized in Step 2 in Figure \ref{MASR2}, with implementation details exemplified in Figure \ref{MASR3}.

Through these multimodal coarse-selection and fine-focusing manipulations, MASR enables the framework to subsequently concentrate more effectively on the portions most relevant to the specific comprehension task.

\subsection{\textbf{Dilated Temporal Expansion (DTE)}}
\begin{figure}
\centering
\includegraphics[width=1.0\linewidth]{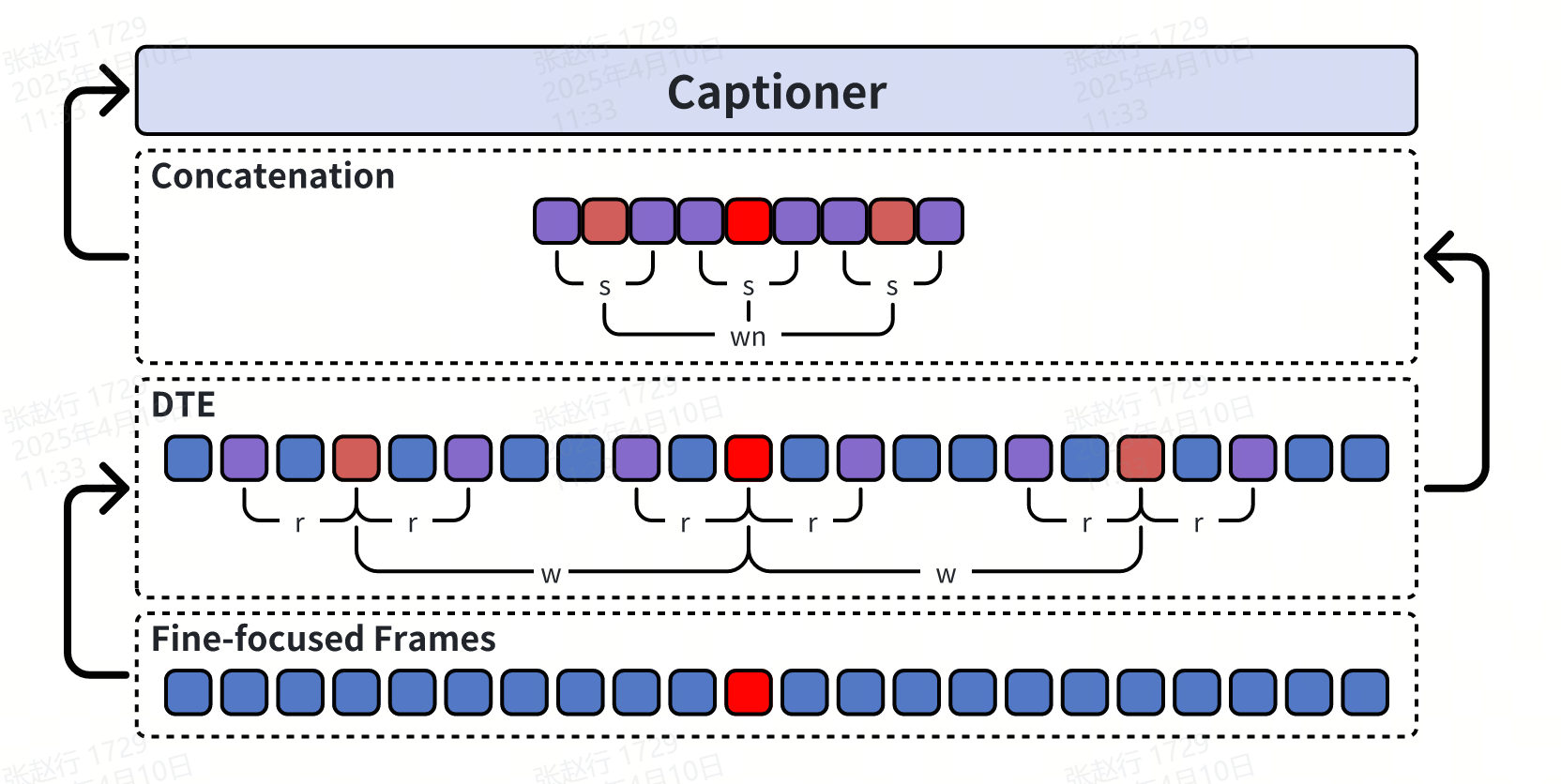}
\captionsetup{justification=raggedright, singlelinecheck=false}
\caption{\label{MASR4}A example of DTE process in Step 3 with the parameters $w=7$, $r=2$, $wn=3$ and $s=3$ showing a total of 9 frames are expanded through DTE for each fine-focused frame. These parameters can be adjusted adaptively.}
\end{figure}

According to the application of expanded receptive fields in dilated convolutional networks for visual detection tasks, we temporally dilate the "temporal receptive field" of each fine-focused frames in
$fcs\_frms$ for broader understanding vision. 
Specifically, based on the expression of 1D convolution $y[n] = \sum_{k=0}^{K-1} x[n + k] \cdot z[k]$, the 1D dilation convolution can be modeled as $y[n] = \sum_{k=0}^{K-1} x[n + r \cdot k] \cdot z[k]$. With the introduction of the dilation rate $r$, the receptive field is thus expanded $r$ times accordingly.
Our DTE takes each focused clip center frame as the anchor and symmetrically selects a total of $wn$ dilated expansion frame scopes in $w$-frame intervals. Within each scope, we then select temporally adjacent frames of each center frame using parameters $s$ and $r$ respectively. Assume that the fine-focused frame index is $n$, its DTE process can be presented as $DTEed\_Frame[n] = \sum_{k=-\lfloor wn/2 \rfloor}^{\lfloor wn/2 \rfloor}\sum_{i=-\lfloor s/2 \rfloor}^{\lfloor s/2 \rfloor}fcs\_frms[n + k \cdot w + i \cdot r]$ where the $\sum$ operand stands for concatenation.
DTE is expected to achieve better video understanding with broader temporal receptive field as illustrated in Figure \ref{MASR4}.

\subsection{\textbf{Iterative Response Confidence Based Self-reflection}}
To avoid the local optima during attention focusing, MASR evaluates the confidence level of each response and uses it as feedback to progressively guide the attention focus of the framework toward the regions more coherent to the query.

Specifically, the LLM in MASR not only generates responses based on acquired context but also evaluates the relevance of extracted information through a confidence score $C$. If $C<=2$, MASR iteratively adjusts attention focusing by repeating Steps 1-4 due to the relevant but inadequate information. When $C=3$, MASR answers directly as it considers the context is enough to generate a confident response.

Unlike other agent-based SOTA solutions like VideoAgent [57] or VCA [50], which introduce an extra reward model to assess confidence scores for each response and sense required supplementary information accordingly, MASR's self-reflection loop enables a single LLM to simultaneously perform three closely related cognitive processes in the semantic space: response generation, evaluation, and relevant clips re-selection as these operations all occur within the same semantic space in our architecture. Our subsequent experiments confirm that this efficient unification does not degrade the system's comprehension capabilities. The complete MASR workflow pseudocode are listed as follows:

\renewcommand{\algorithmicrequire}{\textbf{Input:}}
\renewcommand{\algorithmicensure}{\textbf{Output:}}
\floatname{algorithm}{Algorithm}
\begin{algorithm*}
\caption{The Whole MASR's Algorithm}
\label{alg:Alg1}
    \begin{algorithmic}[1]
        \REQUIRE video $V$, query $Q$, context $P\_ctx$, answer prompt $P\_asw$, selection prompt $P\_slc$, pre-trained vision encoder model $\bold{video\_2\_token(\theta)}$, video frame sampling function $\bold{frame\_sampling(\cdot)}$, video frame cluster function $\bold{frame\_clustering(\cdot)}$, video frame dilated expansion function $\bold{frame\_expansion(\cdot)}$, video frame captioning model $\bold{frame\_captioning(\theta)}$, sematic matching  model $\bold{query\_frame\_matching(\theta)}$ to acquire the most query-relevant frame in the candidate re-focused clips, video clips selection model $\bold{relevant\_clip\_selection(\theta)}$ to acquire the query-relevant clips for confident response, video question answer model $\bold{video\_question\_answer(\theta)}$, maximal allowed repeated times $N$, top relevant visual token candidate number $K_v$ and maximal fine-focused frame number $K_f$, dilated expansion window number $wn$, window interval $w$, frame number per window $s$ and frame interval per window $r$
        \ENSURE final answer $A$ and answering confidence $C$
        \STATE Initialize current repeated time $n \leftarrow 0$, current answering confidence $C \leftarrow 0$, current captions of all the center frames $cfm\_cpts \leftarrow \emptyset$, current focused video clips $fcs\_clps \leftarrow \emptyset$ with their center frames  $fcs\_ccfs \leftarrow \emptyset$
        \STATE Get sampled frames $spl\_frms \leftarrow \bold{frame\_sampling(V)}$
        \STATE Perform clustering to get all the center frames $fcs\_ccfs, fcs\_clps \leftarrow \bold{frame\_clustering(spl\_frms)}$
        \WHILE{$n < N$ and $C < 3$}

            \STATE $epd\_frms \leftarrow \bold{frame\_expansion(fcs\_clps, fcs\_ccfs, wn, w, s, r)}$
            \STATE $cfm\_cpts \leftarrow \bold{frame\_captioning(epd\_frms)}$
            
            \STATE Update $P\_ctx$ with $cfm\_cpts$
            
            \STATE $A, C \leftarrow \bold{video\_question\_answer(Q, P\_asw, P\_ctx)}$

            \IF{$C == 3$}
                \STATE \textbf{break}
            \ENDIF
               
            \STATE $fcs\_clps \leftarrow \bold{relevant\_clip\_selection(Q, P\_slc, P\_ctx, K_c)}$
            \STATE $fcs\_clps, fcs\_ccfs \leftarrow \bold{query\_clip\_matching(\bold{video\_2\_token(fcs\_clps)}, K_f)}$

            \STATE $n=n+1$
            
        \ENDWHILE
    \end{algorithmic}
\end{algorithm*}

\section{Experiments}
\subsection{\textbf{Datasets}}
We conduct comprehensive experiments comparing MASR's performance with other SOTA methods on these video understanding benchmarks:
\begin{itemize}[leftmargin=*]
\item
\textbf{EgoSchema [64]:} It contains 5000 single-choice questions extracted from egocentric videos, with each sample having a duration of 180 seconds. Comprising solely a test set, the dataset includes a subset of 500 questions for which annotated labels are available.
\item
\textbf{NExT-QA [65]:} It comprises 5440 naturalistic videos depicting everyday object interactions, accompanied by 48000 multiple-choice questions. Each video has an average length of 44 seconds. In line with established evaluation protocols, our zero-shot evaluation is performed on the 570 videos that comprise the validation set, accounting for a total of 4,969 tasks.
\item
\textbf{Intent-QA [66]:} This dataset is designed for human intent reasoning with 4,303 videos accompanied by 16000 question-answer pairs. Our evaluation is performed under zero-shot conditions using the test set, concentrating specifically on the 576 necessary videos, which collectively comprise 2,134 tasks. 
\item
\textbf{Video-MME [67]:} The dataset includes diverse ultra-long videos (maximal length over 60 minutes). It takes advantage of the diverse real-world videos and questions requiring spatio-temporal analysis, emotion recognition, and multi-event understanding. 
\end{itemize}

\subsection{\textbf{Implementation Details}}
We evaluate the MASR on all those mentioned datasets under a multiple-choice question answering setup, employing standard accuracy metrics for all experiments.

For EgoSchema [64], IntentQA [66], and NExT-QA [65] datasets, we sample the original videos using 1 FPS while 0.5 FPS for the Video-MME dataset. For EgoSchema [64] and Video-MME [67], we apply the Qwen2-VL-7B [15] model to extract substitles. For IntentQA [66] and NExT-QA [65], we take advantage of LLaVA-NeXT [11] and CogAgent [21] models respectively to generate frame-level captions.

In the comparative experiments, we list other solutions with ChatGPT-4 [7] as the primary model to ensure fairness. Due to ChatGPT-4 [7]'s context length limitation, we configured the DTE parameters as $wn=3$, $s=3$, $r=2$, and $w=6$ for EgoSchema [64], NExT-QA [65], and IntentQA [66] datasets. For Video-MME's [67] long split portions, these parameters are adjusted to $wn=3$, $s=5$, $r=1$, and $w=6$.

During inference, we implement a parallel processing strategy to efficiently extract textual features from concatenated focused frames after DTE, which significantly enhance the inference efficiency.

\subsection{\textbf{Results and Analysis}}
We first test MASR’s performance on four mainstream video datasets mentioned above with various video lengths:
\begin{table*}[htbp]
\centering
\caption{Comparison results on the  EgoSchema, IntentQA, NExT-QA, and Video-MME datasets.} 
\scalebox{0.85}{
\renewcommand{\arraystretch}{1.29}
\begin{tabular}{p{3.5cm}>{\centering\arraybackslash}p{2.5cm}>{\centering\arraybackslash}p{1.2cm}>{\centering\arraybackslash}p{1.2cm}>{\centering\arraybackslash}p{1.cm}>{\centering\arraybackslash}p{1.cm}>{\centering\arraybackslash}p{1.cm}>{\centering\arraybackslash}p{1.cm}>{\centering\arraybackslash}p{1.2cm}}
\midrule
\midrule
\multirow{4}{*}{\textbf{Solutions}} & \multirow{4}{*}{\textbf{(M)LLMs}} & \multicolumn{7}{c}{\textbf{Datasets}} \\ 
\cmidrule{3-9}
 &  & \multicolumn{1}{c}{\textbf{EgoSchema}} & \multicolumn{1}{c}{\textbf{IntentQA}} & \multicolumn{4}{c}{\textbf{NExT-QA}}  & \multicolumn{1}{c}{\textbf{Video-MME}} \\ 
\cmidrule{5-8}
 & & & & \textbf{Temporal} & \textbf{Causal} & \textbf{Descriptive} & \textbf{Average} & \\
\midrule
\multicolumn{9}{l}{\textbf{Based on proprietary MLLMs}} \\
\midrule
LVNet [45] & ChatGPT-4o & \textbf{68.2} & - & 65.5 & 75.0 & 81.5 & 72.9 & - \\
VideoChat2 [32] & ChatGPT-4 & 54.4 & - & - & - & - & 61.7 & 33.2\\
Vamos [73] & ChatGPT-4 & 51.2 & 68.5 & - & - & - & - & -\\
IG-VLM [26] & ChatGPT-4v & 59.8 & 64.2 & 63.6 & 69.8 & 74.7 & 68.6 & -\\
\midrule
\multicolumn{9}{l}{\textbf{Based on open-source MLLMs}} \\
\midrule
MVU [51] & Mistral-13B & 60.3 & - & 55.4 & 48.1 & 64.1 & 55.2 & - \\
LangRepo [54] & Mixtral-8×7B & 66.2 & 59.1 & 51.4 & 64.4 & 69.1 & 60.9 & - \\
SeViLA [25] & BLIP-2 & - & 60.9 & - & - & - & - & -\\
LongVA [76] & Qwen2-7B-Instruct & - & - & - & - & - & - & 46.2\\
InternVL2 [77] & LLaMa & - & - & - & - & - & - & 52.6\\
LLaVa-OneVision-72B [78] & Qwen2 & - & - & - & - & - & - & 60.0\\
Qwen2-VL-72B [15] & Qwen2-VL-72B & - & - & - & - & - & - & \textbf{62.2}\\
\midrule
\multicolumn{9}{l}{\textbf{Based on training-free agents}} \\
\midrule
LLoVi [53] & ChatGPT-4 & 61.2 & 64.0 & 61.0 & 69.5 & 75.6 & 67.7 & 45.4 \\
VideoAgent [58] & ChatGPT-4 & 60.2 & - & 64.5 & 72.7 & 81.1 & 71.3 & 40.2\\
VideoAgent [57] & ChatGPT-4v & 62.8 & - & 60.0 & 76.0 & 76.5 & 70.8 & -\\
GraphVideoAgent [68] & ChatGPT-4 & 62.7 & - & \textbf{74.6} & 65.2 & 83.5 & 73.3 & -\\
LifelongMemory [48]  & ChatGPT-4 & 65.0 & - & - & - & - & 72.3 & -\\
VideoTree [59]   & ChatGPT-4 & 66.2 & 66.9 & 70.6  & \textbf{76.5} & \textbf{83.9} &\textbf{75.6} & 54.2 \\
VideoINSTA [60]    & ChatGPT-4 & 65.0 & \textbf{72.8} & - & - & - & 72.3 & -\\
DrVideo [49]   & ChatGPT-4 & 61.0 & - & - & - & - & - & 51.7 \\
\midrule
CLARF (Ours)    & ChatGPT-4 & \textbf{73.4(+5.2)} & \textbf{73.1(+0.3)} & 70.8 & \textbf{77.2} & \textbf{84.1} & \textbf{75.8(+0.2)} & 57.1\\
\midrule
\midrule
\end{tabular}
}
\label{tab1}
\end{table*}

According to the comparison results summarized in Table \ref{tab1}, MASR outperforms all other SOTA methods (agent-based or video-MLLM-based such as LVNet [45] pre-trained using relevant video datasets) averagely on three datasets. We also list the types of the base models utilized by these methods. On the EgoSchema [64] dataset, it achieves a remarkable 5\% performance gain over the previous leading approach. While on NExT-QA [65] datasets, MASR gives better overall performance compared to other methods by 0.2\%. Across its four sub-categories, MASR also achieves Top-1 rankings in three categories and earns the second place in the rest one. For the Intent-QA [66] dataset, MASR surpasses the second method by a margin of 0.2\%. These accomplishments serve as compelling evidence for the effectiveness of our proposed method in handling video comprehension tasks.

To further highlight the advantages of our method in processing long-length videos, we also conduct the challenging comparison on the long split part of the Video-MME [67]. As shown in Table \ref{tab1} again, our method achieves 57.1\% in response accuracy, outperforming all the other listed SOTA agent-based solutions and some fine-tuning-required open-source video models like InternVL2 [77]. Noticeably, those solutions with better performance than ours utilize the update-to-dated massive models with much larger parameter sizes and higher training cost than our core models. 

\begin{figure}
\centering
\includegraphics[width=1.0\linewidth]{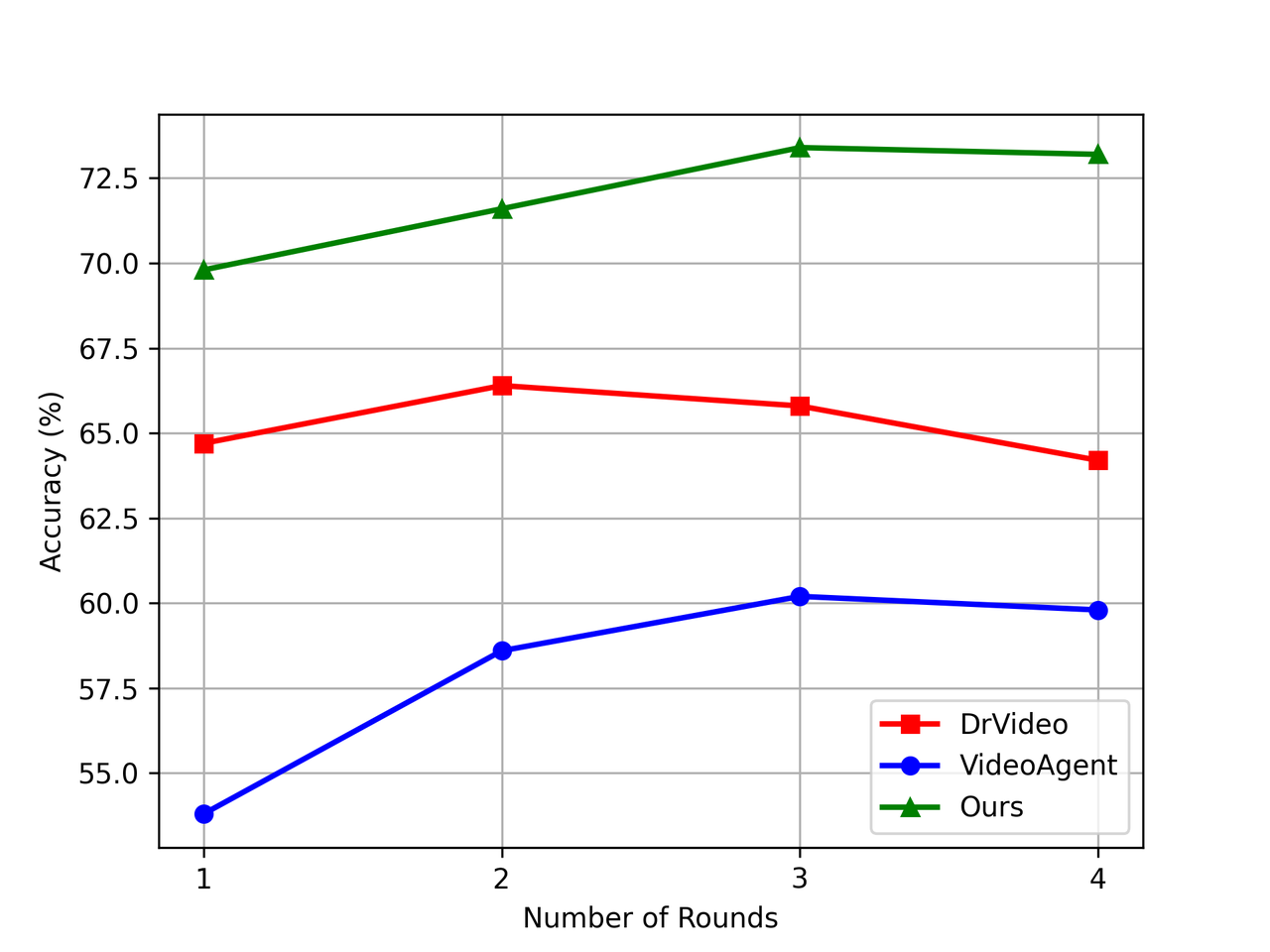}
\captionsetup{justification=raggedright, singlelinecheck=false}
\caption{\label{MASR5}A demonstration of MASR's reflective reasoning processes to answer questions in the EgoSchema dataset.}
\end{figure}

We also investigate the impact of the self-reflection mechanism on the accuracy. Figure \ref{MASR5} displays the comparison between two other self-reflection-powered agent-based solutions and our MASR on the EgoSchema [64] dataset under different self-reflective rounds. We find that:

MASR significantly improves response accuracy through multi-round self-reflection, consistently outperforming DrVideo [49] and VideoAgent [57]. While the accuracy of DrVideo [49] peaks in the second round and VideoAgent [57] at the third round, our MASR achieves stable precision increase from more rounds, highlighting the effectiveness of its self-reflection mechanism. In addition, DrVideo [49] suffers from overthinking as the response accuracy decreases, showing that more comprehension through excessive information only does not necessarily benefit the model. 

\subsection{\textbf{Case Study}}
\begin{figure*}
\centering
\includegraphics[width=1.0\linewidth]{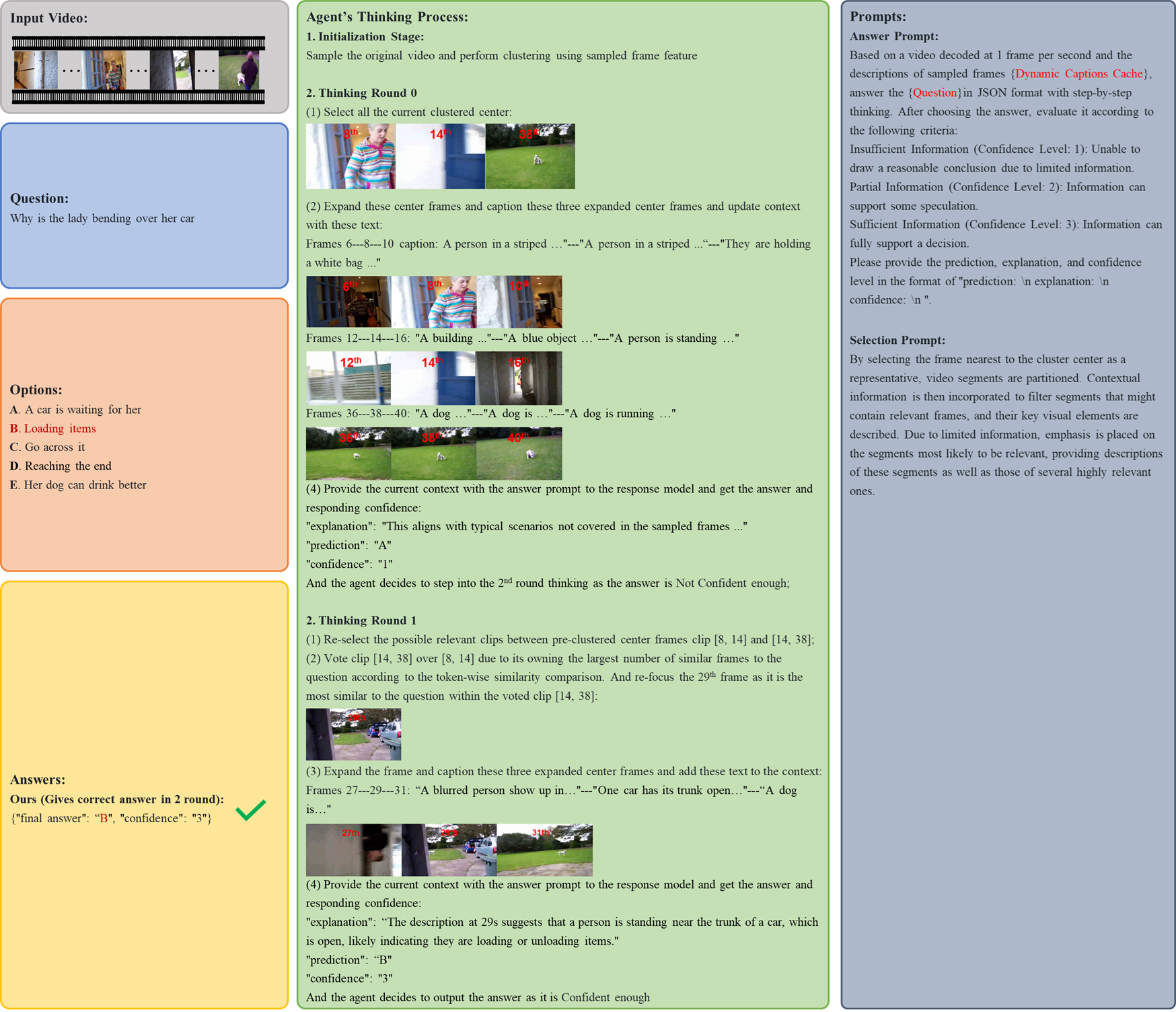}
\captionsetup{justification=raggedright, singlelinecheck=false}
\caption{\label{MASR6}A demonstration of MASR's reflective reasoning processes to answer questions in the EgoSchema [64] dataset.}
\end{figure*}

\begin{figure*}
\centering
\includegraphics[width=1.0\linewidth]{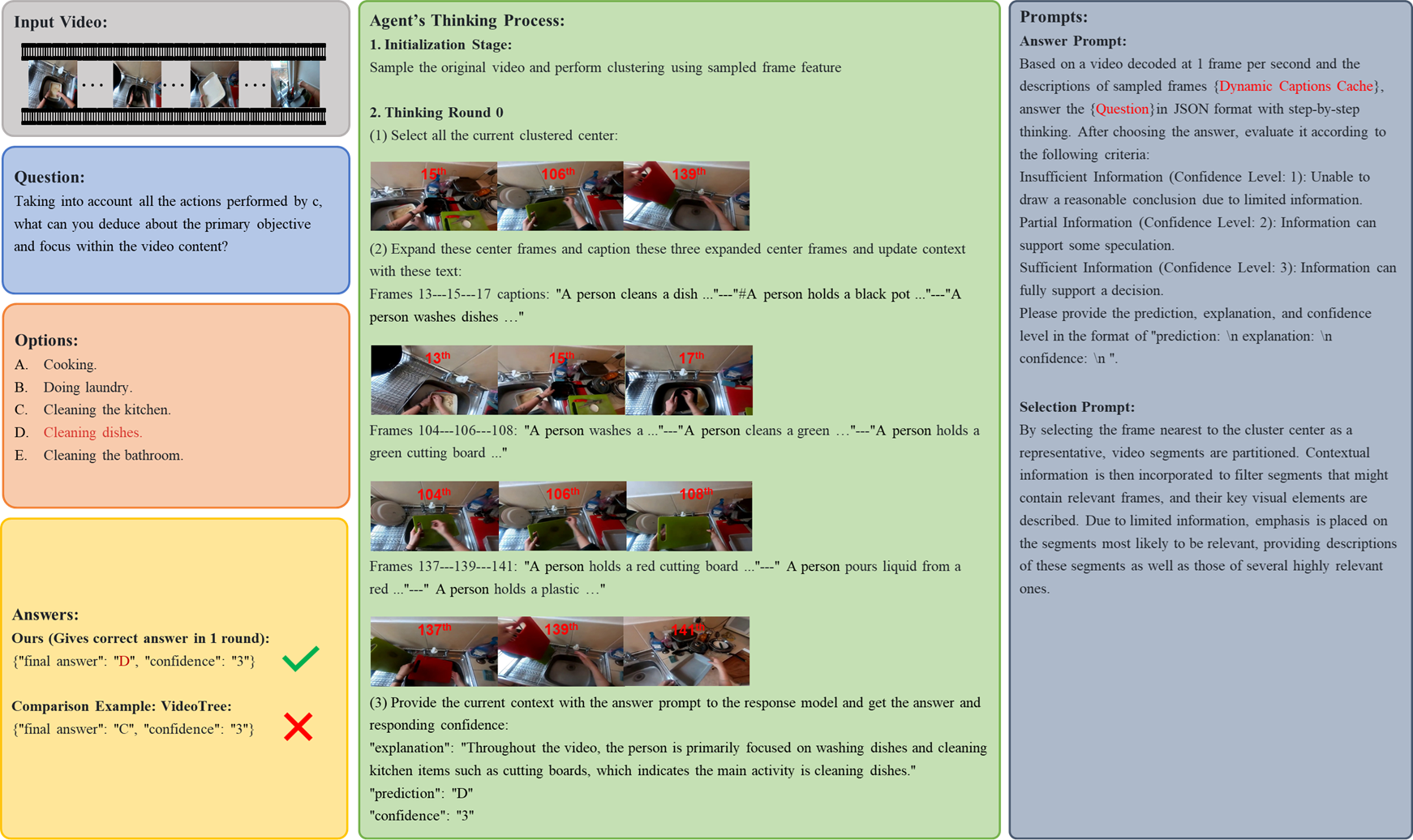}
\captionsetup{justification=raggedright, singlelinecheck=false}
\caption{\label{MASR7}A demonstration of MASR's reflective reasoning processes to answer questions in the IntentQA [66] dataset.}
\end{figure*}

Figures \ref{MASR6} and \ref{MASR7} present MASR's reflective reasoning processes from the EgoSchema [64] and IntentQA [66] datasets,respectively. Initially, we define a round as complete only after executing the operations of the relevant clips selection model; otherwise, it is marked as round zero.In Figure \ref{MASR6}, MASR first samples the original video at 1 FPS and obtains three cluster center: the 8th, the 14th and the 38th frame. MASR directly makes these three clustered center frames as the fine-focused frames and performs DTE on them with parameters $wn=1$, $w=0$, $r=2$, and $s=3$ before the captioning-based semantic extraction. Since MASR cannot produce a high-confidence answer based on the this context, it proceeds to coarsely select the clip from the 14th to the 38th frame as the coarse selection result. Subsequently, through fine-focusing, it senses the 29th frame as the most query-relevant to update the context. This enhanced context is able to provide MASR with more relevant and comprehensive information and successfully predict a highly confident and correct answer. Similarly, Figure \ref{MASR7} illustrates MACF's reasoning process for another test case from the IntentQA [66] dataset, which is otherwise not correctly responded in VideoTree [59]'s framework.

\subsection{\textbf{Ablation Experiments}}
We design well-rounded ablation experiments on the EgoSchema [64] dataset, which yielded the best experiment results, to demonstrate the significance of the main modules in MASR.

We first perform the ablation comparison to evaluate two key creative modules in the MCRS framework: the MCRS, the DTE and the self-reflection modules. From Table \ref{tab2}, around 8. 1\% of all queries cannot be answered correctly in a single round without self-reflective feedback. This shows the excellency of our self-reflective mechanism as a whole in addressing questions that are difficult to answer correctly on the first attempt. We further perform the ablation comparison on the key module MCRS. We use the token-based similarity matching instead of MCRS, resulting in a remarkable 7.4\% decrease in accuracy. Especially in cases of queries to summarize or generalize, relying solely on token-wise similarity matching is highly possible to lead to incorrect answers because it confines the attention focus to local patterns. 
The significance of DTE is also demonstrated in Table \ref{tab2} by its 9.4\% gain in accuracy as expected. It proves that efficient semantic information extraction is also indispensable on top of the exquisitely designed attention-focused mechanism. 
\begin{table}[htbp]
\centering
\caption{Ablation experiment on MASR's core components} 
\scalebox{1.}
{
    \renewcommand{\arraystretch}{1.29}
    \begin{tabular}{l|c} 
        \midrule
        \midrule
        \textbf{Condition} & \textbf{Accuracy} \\
        \midrule
        Complete MASR & 73.4 \\
        \midrule
        w/o Self-Reflection & 65.3(-8.1) \\
        w/o MCRS & 66.0(-7.4) \\
        w/o DTE & 64.1(-9.3) \\
        \midrule
        \midrule
    \end{tabular}
}
\label{tab2}
\end{table}

Table \ref{tab3} displays the impact of different visual encoders in the fine-level focusing step of the MCRS module on the response accuracy. The comparison reveals that the parameter size and the input image resolution of the visual encoder have a positive impact on overall accuracy. 
\begin{table}[htbp]
\centering
\caption{Ablation experiment on visual encoders} 
\scalebox{0.9}
{
    \renewcommand{\arraystretch}{1.29}
    \begin{tabular}{l|c|c|c} 
        \midrule
        \midrule
        \textbf{Visual Encoder} & \textbf{Parameters} & \textbf{Resolution} & \textbf{Accuracy} \\
        \midrule
        OpenCLIP-ViT-G [72] & 1B & 224 & 70.5 \\
        \textbf{EVA-CLIP-8B [69]} & 8B & 224 & \textbf{73.4} \\
        EVA-CLIP-8B-plus [69] & 8B & 448 & 71.4 \\
        \midrule
        \midrule
    \end{tabular}
}
\label{tab3}
\end{table}

To evaluate the VLM's semantic extraction capability in MASR, we incorporate three VLM-based captioners: frame-based Qwen-2-VL-7B [63], LLaVA-NeXT [11], and clip-based LaViLa [62]. As shown in Table \ref{tab4}, Qwen-2-VL-7B achieves the best performance due to its superiority in capturing dynamic object information, which is better suited to the content of our EgoSchema [64] dataset.
\begin{table}[htbp]
\centering
\caption{Ablation experiment on VLMs as captioners} 
\scalebox{1.}
{
    \renewcommand{\arraystretch}{1.29}
    \begin{tabular}{l|c|c} 
        \midrule
        \midrule
        \textbf{Captioner} & \textbf{Input} & \textbf{Accuracy} \\
        \midrule
        LLaVA-NeXT [11] & Frame-wise & 68.1 \\
        \textbf{Qwen2-VL-7B} [15] & Frame-wise & \textbf{73.4} \\
        LaViLa [62] & Clip-wise & 71.1 \\
        \midrule
        \midrule
    \end{tabular}
}
\label{tab4}
\end{table}

To assess the ability of single-LLM-based reflector in our MASR, Table \ref{tab5} demonstrates the comparison among these three integrated LLMs. Since the LLM in MASR needs to play the roles of the responser, evaluator and coarse relevant video clip selector, models with strong reasoning power such as DeepSeek-V3 [3] and ChatGPT-4 [7] achieve higher accuracy as expected. As the reasoning ability of the LLM improves, the response accuracy of the MASR framework also increases.
\begin{table}[htbp]
\centering
\caption{Ablation experiment on LLMs as reflectors} 
\scalebox{1.}
{
    \renewcommand{\arraystretch}{1.29}
    \begin{tabular}{l|c|c|c} 
        \midrule
        \midrule
        \textbf{Reflector} & \textbf{Type} & \textbf{Size} & \textbf{Accuracy} \\
        \midrule
        Llama-3.3-70B [1] & Open & 70B & 66.9 \\
        DeepSeek-V3 [3] & Open & 70B & 70.6 \\
        Qwen2.5-72B [5] & Open & 72B & 69.7 \\
        \textbf{ChatGPT-4 [7]} & Proprietary & - & \textbf{73.4} \\
        \midrule
        \midrule
    \end{tabular}
}
\label{tab5}
\end{table}

Table \ref{tab6} presents the effective of the relevant frame candidate number $K_v$ in the fine-focusing step in the MCRS module. Ignoring the computation cost, the participation of more frames from coarse-selected clip candidates is able to generate more relevant fine-level match. The results validate this expectation.
\begin{table}[htbp]
\centering
\caption{Ablation experiment on number of frame candidate to perform fine-focusing} 
\scalebox{1.}
{
    \renewcommand{\arraystretch}{1.29}
    \begin{tabular}{c|c} 
        \midrule
        \midrule
        \textbf{Similarity Candidates} & \textbf{Accuracy} \\
        \midrule
        30 & 70.6 \\
        60 & 73.0 \\
        \textbf{90} & \textbf{73.4} \\
        \midrule
        \midrule
    \end{tabular}
}
\label{tab6}
\end{table}

Table \ref{tab7} and \ref{tab8} illustrate the relation between the DTE's expansion extent and the degree response accuracy in the MASR framework. We list $wn$ and $r$ in our ablation experiments, which are the key hyperparameters in our framework. Other parameters remain fixed. The results show that higher expansion does not necessarily lead to higher accuracy, as excessive temporal expansion may introduce noise into the previously attention-focused frames.
\begin{table}[htbp]
\centering
\caption{Ablation experiment on DTE's scope} 
\scalebox{1.}
{
    \renewcommand{\arraystretch}{1.}
    \begin{tabular}{c|c} 
        \midrule
        \midrule
        \textbf{$wn$ in DTE} & \textbf{Accuracy} \\
        \midrule
        1 & 69.0 \\
        \textbf{3} & \textbf{73.4} \\
        5 & 71.2 \\
        \midrule
        \midrule
    \end{tabular}
}
\label{tab7}
\end{table}
\begin{table}[htbp]
\centering
\caption{Ablation experiment on DTE's frame intervals} 
\scalebox{1.}
{
    \renewcommand{\arraystretch}{1.}
    \begin{tabular}{c|c} 
        \midrule
        \midrule
        \textbf{$r$ in DTE} & \textbf{Accuracy} \\
        \midrule
        1 & 70.4 \\
        \textbf{2} & \textbf{73.4} \\
        3 & 71.4 \\
        \midrule
        \midrule
    \end{tabular}
}
\label{tab8}
\end{table}

\section{Future Works}
In this work, we propose MASR, an efficient agent-based video understanding framework. It features multimodal coarse-to-fine relevance sensing and enhanced dilated temporal expansion for semantic extraction, organized by human-like iterative self-reflection. We demonstrate that our MASR achieves state-of-the-art (SOTA) performance and efficiency, without the need for heavy supervised fine-tuning with in-domain data or other customized tools. 

Despite MASR's SOTA performance on the tested datasets, it still faces the following challenges: (1) High computational latency is considered a primary bottleneck during long-video processing; this problem also exists in other concurrent solutions like LLoVi [53] and DrVideo [49]. (2) Achieving an adaptive balance between the retention and removal of semantic information in context during each self-reflection round. In the future, taking advantage of a wider range of datasets or more advanced analytical methods could yield deeper insights and further strengthen the applicability of the results.



\end{document}